\documentclass[letterpaper, 10 pt, conference]{ieeeconf}  

\IEEEoverridecommandlockouts                              

\usepackage{amsmath} 
\usepackage{graphicx}
\usepackage{tabularx}
\usepackage{booktabs}
\usepackage{balance}

\title{\LARGE \bf
CognitiveReality: Robot-Agnostic Semantic Gaussian Mapping\\with an LLM Agent for Immersive Collaborative VR Teleoperation
}

\author{Timofei Kozlov$^{*1}$, Dmitrii Maliukov$^{*1}$, Andrey Marchenko$^{2}$, \\
Dmitrii Plotnikov$^{1}$, Miguel Altamirano Cabrera$^{1}$, and Dzmitry Tsetserukou$^{1}$
\thanks{$^{*}$These authors contributed equally to this work.}
\thanks{$^{1}$Intelligent Space Robotics Laboratory, Skolkovo Institute of Science and Technology, Moscow, Russian Federation. {\tt\small {\{Timofei.Kozlov, Dmitrii.Maliukov, Dmitrii.Plotnikov2 M.Altamirano, D.Tsetserukou\}} @skoltech.ru  }
$^{2}$ NLP Research Center, Moscow, Russian Federation {\tt\small marchenko.ai@phystech.edu}}
 }

\begin{document}

\bstctlcite{IEEEexample:BSTcontrol}
\maketitle
\thispagestyle{empty}
\pagestyle{empty}

\begin{abstract}
 A photorealistic 3D view tells a teleoperator where a robot is, but not what the scene contains, how well each object has been observed, or how to turn pointing and speech into robot action. CognitiveReality turns a robot's RGB-D stream into a live, semantically indexed Gaussian-TSDF map shared by an operator in virtual reality and a tool-using language agent. One mapper binary serves any platform through configuration alone: it ingests poses from robot SLAM, joint kinematics, motion capture or an inline visual tracker, bridges localization outages with a shadow tracker and keyframe-anchored PnP, and maintains open-vocabulary instance identities with per-object quality at 2 Hz. Speech and controller rays are grounded against persistent scene objects through validated typed tools and operator-confirmed robot actions. In the controlled agent evaluation, the deployed local Qwen3-VL-8B router reaches 81.24\% tool exact match, while merge-aware replay correctly redirects 101 absorbed object identifiers. On robot data CognitiveReality exceeds a Gaussian-plus-SDF baseline by 2-8 dB; pose error through 5-40 s SLAM outages stays within 1-8 cm. Deployed live on two quadrupeds, the agent executed 26 of 30 navigation requests and 20 of 20 re-observation requests, raising object quality by 2-5 dB.

\end{abstract}

\section{INTRODUCTION}

Teleoperation interfaces for mobile robots increasingly replace flat camera feeds with 3D reconstructions shown in a head-mounted display. A high-quality photorealistic representation can provide a more interpretable and natural view of the scene, enabling a more intuitive and immersive operator experience. For this reason, 3D Gaussian Splatting (3DGS) has recently gained attention as a promising approach for photorealistic scene reconstruction and visualization. However, even high-quality photorealistic reconstructions remain primarily passive representations of the environment and do not inherently provide users with an intuitive way to understand the scene, refer to its contents, or interact with a robot operating within it.

The key interface is therefore not the rendered appearance alone, but the
object-level state exposed by the evolving map. Persistent identities, geometry, semantic labels, and reconstruction-quality attributes allow the
operator and agent to refer to the same scene entities, while typed actions and validation separate language interpretation from execution. We evaluate this
shared Scene Model both as an agent-facing interface and as the basis of a live operator-agent-robot loop.

\begin{figure}
    \centering
    \includegraphics[width=0.8\linewidth]{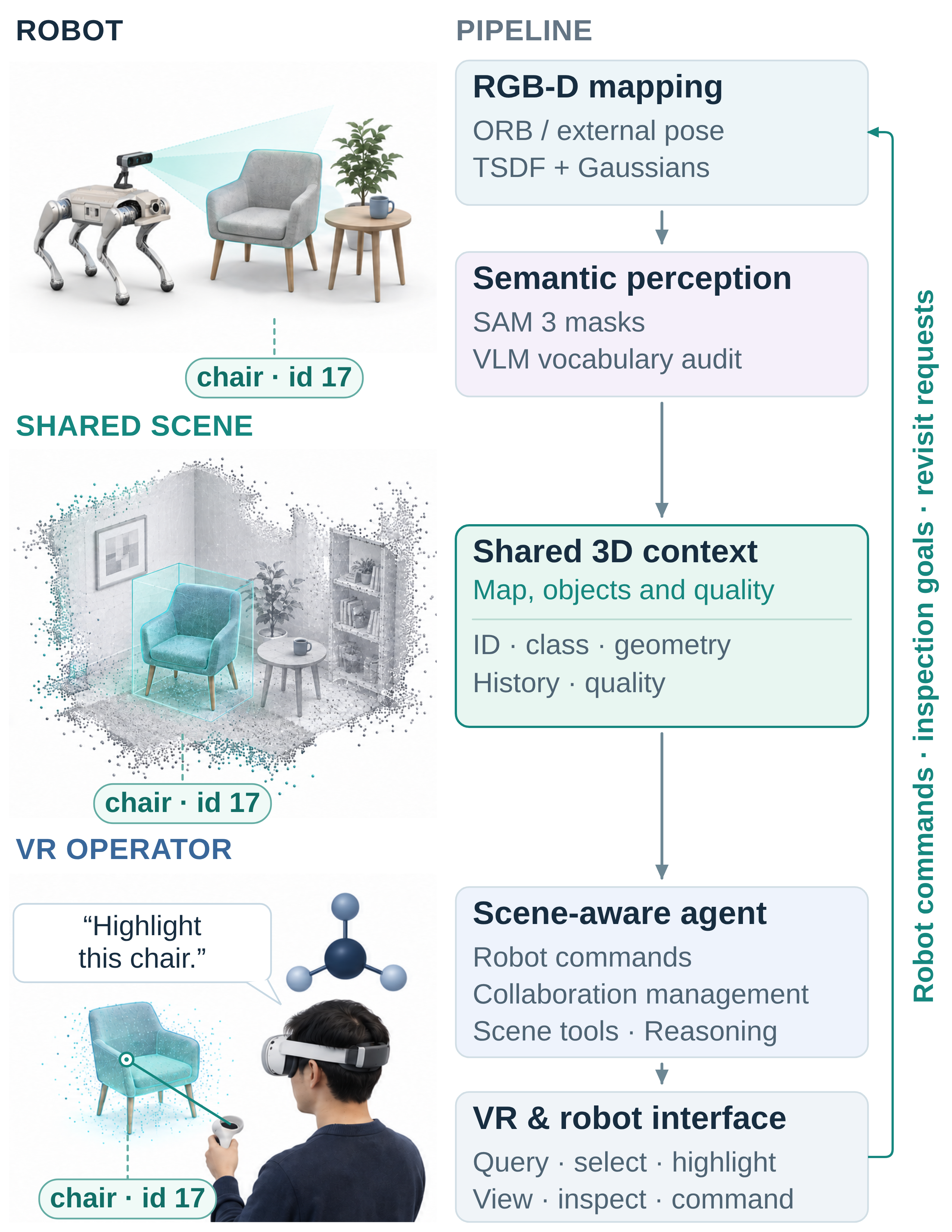}
    \caption{Shared 3D scene memory for robot-agent-VR interaction. RGB-D observations from a robot build an online TSDF-Gaussian map enriched with semantic labels, object identities, and observation quality. The agent grounds the operator’s language and pointing in this shared memory, enabling scene queries, object selection, highlighting, and viewpoint control. A common object ID connects robot perception, scene reconstruction, and VR interaction.}
    \label{fig:title}
\end{figure}

In this work, we introduce a real-time SLAM and interaction system that combines high-fidelity 3D Gaussian and SDF-based scene reconstruction with a VLM-powered agent for intuitive human-robot interaction (Fig.~\ref{fig:title}). The reconstruction backend provides a detailed and photorealistic representation of the environment in real time, while the VLM-powered agent enables users to interact with the reconstructed scene and a robot operating within it through natural language and high-level commands. Rather than limiting the reconstructed scene to visualization, the agent provides an accessible interface for scene understanding, object-level interaction, and robot control. To further enhance user interaction, we implemented a dedicated VR rendering pipeline that enables immersive real-time visualization of the reconstructed environment and agent-mediated interaction with both the scene and the robot.

Our contributions are:

1. \textbf{Robot-agnostic online Gaussian-SDF mapping.} The mapper takes its poses from either the robot's SLAM system or an inline visual tracker (ORB-SLAM3) tuned for fast and robust tracking, and switches between the two at run time. Supervision frames may be filtered using the camera's or robot's gyroscope for blur minimization, and depth fusion is weighted by a sensor noise model. On robot data the mapper exceeds the Gaussian-plus-SDF~\cite{peng2025gpsslam} baseline by 2-8 dB.

2. \textbf{Map-continuous relocalization through localization outages.} A shadow visual tracker bridges outages of the robot's localization, and keyframe-anchored PnP bridges a simultaneous loss of the visual tracker. Pose error through 5-40 s outages (both robot's SLAM and our shadow tracker) stays within 1-8 cm, and the map rendered inside the outage window is indistinguishable from an uninterrupted reference.

3. \textbf{Online instance semantics and a Scene Model.} An open-vocabulary segmenter labels the map while it is being built. Instance identities are stored on the Gaussians themselves and persist as the map grows. Semantics run at 2 Hz beside the mapper on one consumer GPU without dropping mapper frames. The result is exported as a Scene Model: a list of objects with class, position, coverage and per-object rendering quality that the agent and the operator use during the mission.

4. \textbf{A scene-aware agent with typed tools.} Speech and pointing are
resolved against the live Scene Model through tools for inventory, semantic
search, description, highlighting, quality reports, memory, robot goals, and
inspection missions. A local VLM produces structured calls that are checked by
pre-routing, scene validation, and operator confirmation. We evaluate this
interface through 1,471 command cases, temporal reference replay, metadata
controls, and live robot execution.

5. \textbf{An immersive operator interface and live deployments.} The system supports controller-ray selection of instances/floor goals, ray-grounded push-to-talk, and route, trail, highlight, and quality overlays. These are added to a 90 Hz host-rendered stereo stream featuring a 3D robot model that is either animated or mirrors physical joint movements. The system is evaluated on several robotic platforms with and without onboard SLAM and the loop from headset pointing to robot's movement and map expansion is demonstrated.

\textbf{To facilitate reproducibility and future research, our complete source code and datasets will be made publicly available upon acceptance.}

\section{RELATED WORK}

\subsection{Online Reconstruction and Semantic Mapping}

3D Gaussian Splatting enables photorealistic scene reconstruction~\cite{kerbl2023gaussian}, while hybrid Gaussian-SDF methods such as GPS-SLAM improve online geometric fusion and Gaussian initialization. CognitiveReality (CR) follows this design but supports interchangeable robot localization and inline ORB-SLAM3~\cite{orbslam3}, vibration-gated supervision, uncertainty-weighted depth fusion, and localization-outage recovery.

Open-vocabulary 3D mapping has been explored with object graphs such as ConceptGraphs~\cite{gu2024conceptgraphs} and Gaussian representations. LangSplat~\cite{qin2024langsplat} distills language features into Gaussians, Gaussian Grouping~\cite{ye2024gaussiangrouping} embeds instance identities, and online approaches include OpenGS-SLAM~\cite{yang2025opengsslam}, Online Language Splatting~\cite{katragadda2025onlinelang}, and OnlinePG~\cite{zhai2026onlinepg}. In contrast, our system maintains persistent object identities and reconstruction-quality attributes in a live Scene Model shared by the robot, agent, and operator.

\subsection{Gaussian Scenes and Immersive Teleoperation}

RFFR~\cite{wildersmith2024rffr} combines online radiance-field reconstruction with ROS and VR, Boehringer et al.~\cite{boehringer2025immersive} use Gaussian scenes for immersive mobile manipulation, and HIL-GS~\cite{lee2026hilgs} incorporates operator-guided viewpoints into reconstruction. VRSplat~\cite{tu2025vrsplat} supports high-performance Gaussian VR rendering, while GaussAnything~\cite{maliukov2026gaussanything} performs incremental Gaussian rendering directly on standalone headsets. CognitiveReality supports both remote and headset-local rendering while preserving common object IDs for visualization, language grounding, highlighting, and robot actions.

\subsection{Tool Use and Multimodal Human-Robot Interaction}

Typed action interfaces connect language reasoning to executable skills or environment feedback~\cite{saycan,react}, where action selection remains distinct from constructing valid, argument-complete calls~\cite{bfcl}. Multimodal interfaces using speech and pointing establish early paradigms for object selection~\cite{bolt}. Our system integrates these concepts with a versioned semantic map and confirmation gates, evaluating complete-call correctness, sandbox execution, object grounding, and reference validity post-map-updates.

\subsection{Agent-World Interaction}

LLM-based planning leverages structured 3D scene graphs in SayPlan~\cite{rana2023sayplan} and ConceptGraphs~\cite{gu2024conceptgraphs}. Agentic Scene Policies~\cite{morin2025asp} exposes object-centric maps through typed tools for retrieval, interaction, and navigation. Over Gaussian maps, GSMem~\cite{lu2026gsmem} uses the representation as persistent spatial memory with rendered views for VLM-based reasoning, while 3DGSNav~\cite{zheng2026gsnav} uses renders of an incrementally built map for frontier selection. VISTA~\cite{nagami2025vista} further explores online semantic Gaussian mapping for task-relevant robot exploration. Rendering Gaussian maps for VLMs is thus an established pattern.

CognitiveReality distinguishes itself by anchoring this workflow in a live, robot-constructed map viewed from the operator's perspective. Our object model incorporates coverage and rendering quality for agent queries, routing all actions through validated typed tools~\cite{royce2025rosa}. To our knowledge, this is the first system to unify a live robot-built Gaussian map, a tool-using agent, and an operator grounding commands by pointing into a shared map.

\section{SYSTEM ARCHITECTURE}

\begin{figure*}[t]
\vspace{8pt}
    \centering
    \includegraphics[width=\textwidth]{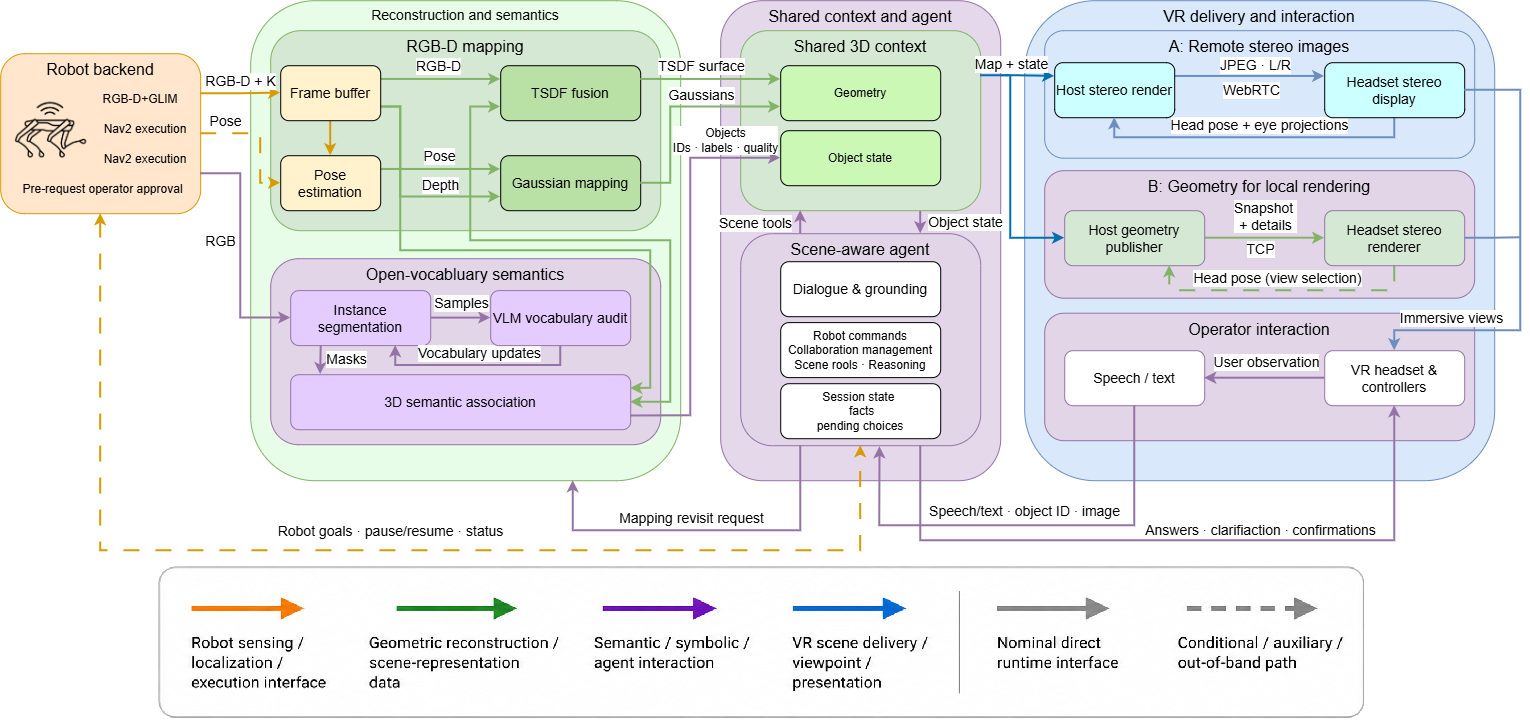}
    \caption{Architecture of CognitiveReality, showing the data flow between the robot backend, online geometric and semantic reconstruction, the shared scene-aware agent, and the two VR delivery modes. A common 3D scene representation connects robot sensing, semantic object state, agent reasoning, operator interaction, and robot control.}
    \label{fig:architecture}
    \vspace{-4mm}
\end{figure*}

\subsection{Online Tracking and Reconstruction}

We build on the Gaussian-SDF design of GPS-SLAM: depth is fused into a signed distance field, which is stable under sensor noise and cheap to update, while 3D Gaussians optimized over the fused surface provide photorealistic rendering that the SDF alone cannot. To mitigate noise from raw RGB-D measurements, Gaussian initialization is jointly guided by photometric errors and geometric consistency. Candidate pixels are filtered using raycast depth and TSDF confidence metrics. Reliable candidates are then sampled and initialized near the reconstructed surface, anchoring Gaussian growth to the underlying geometry. Gaussian parameters are optimized periodically and asynchronously using recent frames and selected keyframes, ensuring the compute-intensive optimization loop does not block real-time frame acquisition. Geometrically inconsistent or poorly constrained primitives are filtered out via subsequent pruning steps.

 The resulting dual representation evolves continuously without requiring offline training, enabling interactive querying at any timestamp. For evaluation, novel views are rendered via the standard Gaussian rasterization pipeline. For immersive visualization, stereo views are rendered based on the user's VR headset pose and streamed via WebRTC, providing the user with immediate, interactive feedback of the live reconstruction process. Alternatively, the scene is exported as a simplified model to be rendered on a headset, adapting the GaussAnything methodology.

The simplified Scene Model is created every 2 seconds in form of a JSON file with Scene geometric size, number of Gaussians, all objects with their classes and IDs, the centroids and bounding boxes of these objects as well as their reconstruction quality, and some other Scene parameters.

 \textit{Map-continuous relocalization.} Our framework treats the robot's primary localization as a replaceable pose source, backed by a secondary visual tracker that estimates the transformation to the global map frame. During primary tracking outages, this visual alignment serves as a redundant pose layer. If both sources fail, incoming frames are registered against the three nearest RGB-D keyframes using PnP-RANSAC. High-confidence registrations are initialized as new keyframes, extending the tracking chain into unmapped spaces. Because the system relies on stored keyframes as anchors rather than map renderings, the global map is never reset, ensuring a seamless handover once the external pose source recovers.
 
\begin{figure}[t]
\vspace{8pt}
  \centering
  \includegraphics[width=1.0\linewidth]{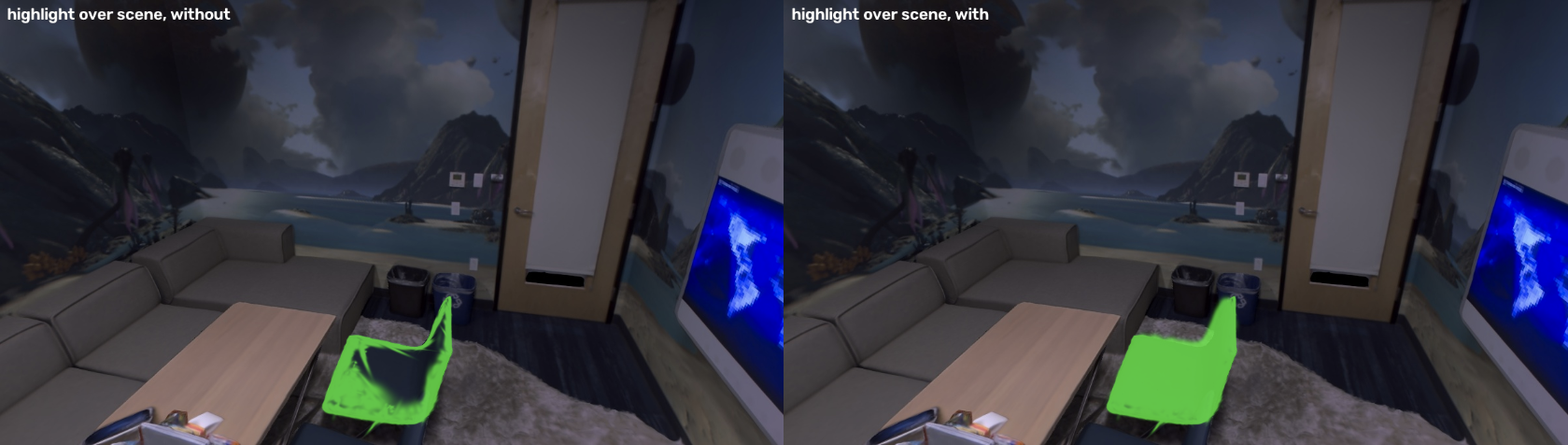}
  \caption{Example of the semantic coverage without (left) and with (right) Label Splats on a chair in the Replica office0 scene.}
  \vspace{-0.6cm}
  \label{fig:label_splats}
\end{figure}

\textit{Label Splats.} To address incomplete object coverage in hybrid representations, where appearance Gaussians are seeded exclusively in high-error regions, we introduce frozen, semantic-only Gaussians - Label Splats. Semantic pixels that exhibit geometric consistency across two independent views are placed directly on the SDF surface. These primitives carry semantic identifiers and votes for object centering, but are excluded from the photometric rasterizer to preserve rendering fidelity. This approach achieves \(100\%\) semantic coverage on the Replica office0 scene (vs 47.6\% without Label Splats), correcting object centroids toward their true geometric centers to enable accurate approach and viewpoint planning as well as consistent display when highlighting (Fig.~\ref{fig:label_splats}). Maintaining all semantic identifiers on a unified primitive layer provides a single, efficient lookup interface for downstream tasks.

\textit{Semantic VLM auditor.} Separate from the interaction router, the semantic auditor maintains the segmenter's vocabulary as a dynamic session state rather than relying on frame-level or one-shot prompting. We propose an asynchronous dual-rate architecture. A fast tracking loop runs at 2-3 Hz, while a slow discovery loop operates at 0.5 Hz (median latency: 0.6 s), triggered when unlabeled regions appear. The slow loop leverages a local vision-language model to verify existing masks and identify novel objects. To ensure robustness, new open-world classes can be filtered through an independent viewpoint confirmation rule, and synonyms are unified via a text-embedding gate (SigLIP + difflib). Stale classes are automatically retired. Consequently, novel objects are registered within 1-2 s without the confirmation rule, or 4-5 s when enforcing a strict two-view confirmation constraint.

\subsection{VR Rendering and Streaming}

We support two VR rendering modes: \textbf{remote stereo image streaming (RSIS)} and \textbf{simplified scene streaming with local rendering (3SLR).} 

In the \textbf{RSIS mode}, the host renders the left- and right-eye images. The headset sends its pose and eye projection matrices. The renderer combines TSDF raycasting with Gaussian appearance and object highlighting. Tracking data and JPEG-compressed eye images are sent through WebRTC data channels.

This mode reduces the rendering and memory load on the headset. However, head-motion response depends on network delay, host rendering, image encoding, and frame delivery.

In the \textbf{3SLR mode}, the host sends a TSDF mesh and quantized Gaussian primitives. The headset stores this data and renders both eyes locally using OpenXR and OpenGL ES. After the first full scene transfer, the host sends only added, changed, or removed Gaussians and updated meshes. Persistent IDs link Gaussian updates and object highlighting~\cite{maliukov2026gaussanything}.

This mode allows fast viewpoint changes between network updates, but it requires more headset memory and computing power. Visual quality also depends on the size and age of the local scene cache.

\subsection{Human-agent interaction}

The proposed human-agent framework provides a multimodal interface for collaborative scene exploration and manipulation. The operator communicates via speech, text, object selection, or spatial pointing, optionally supplemented by the user's VR viewport image. Interactions are grounded in a shared 3D scene model containing semantic labels, spatial boundaries, and reconstruction-quality metrics. Spatial references (e.g., "this chair") are resolved by casting pointing rays onto scene objects, mapping them to persistent object IDs, and blending them with the dialogue context $h$. A language-model-based router interprets the request to execute scene operations, including object queries, visual question answering via a VLM, viewpoint control, and spatial annotations.

To decouple language interpretation from execution, the agent operates on a versioned symbolic scene model rather than the raw Gaussian representation. This model exposes persistent integer IDs, semantic labels with confidences, centroids, 3D bounding boxes, and reconstruction attributes, serving as a unified interface between reconstruction, language grounding, and robot missions.

Formally, each user interaction is represented as an input tuple $\mathbf{x} = (u, g, s, h)$, where $u$ denotes the text or speech request, $g \in \{\emptyset, \text{point}, \text{select}\}$ represents the spatial grounding, $s$ is the scene state, and $h$ is the interaction history. 
The agent outputs either a clarification request or a typed decision $\mathbf{d} = (k, n, a)$, comprising a decision kind $k$, a tool or mission name $n$, and structured arguments $a$.

 \begin{figure}[t]
    \vspace{8pt}
     \centering
     \includegraphics[width=0.6\columnwidth]{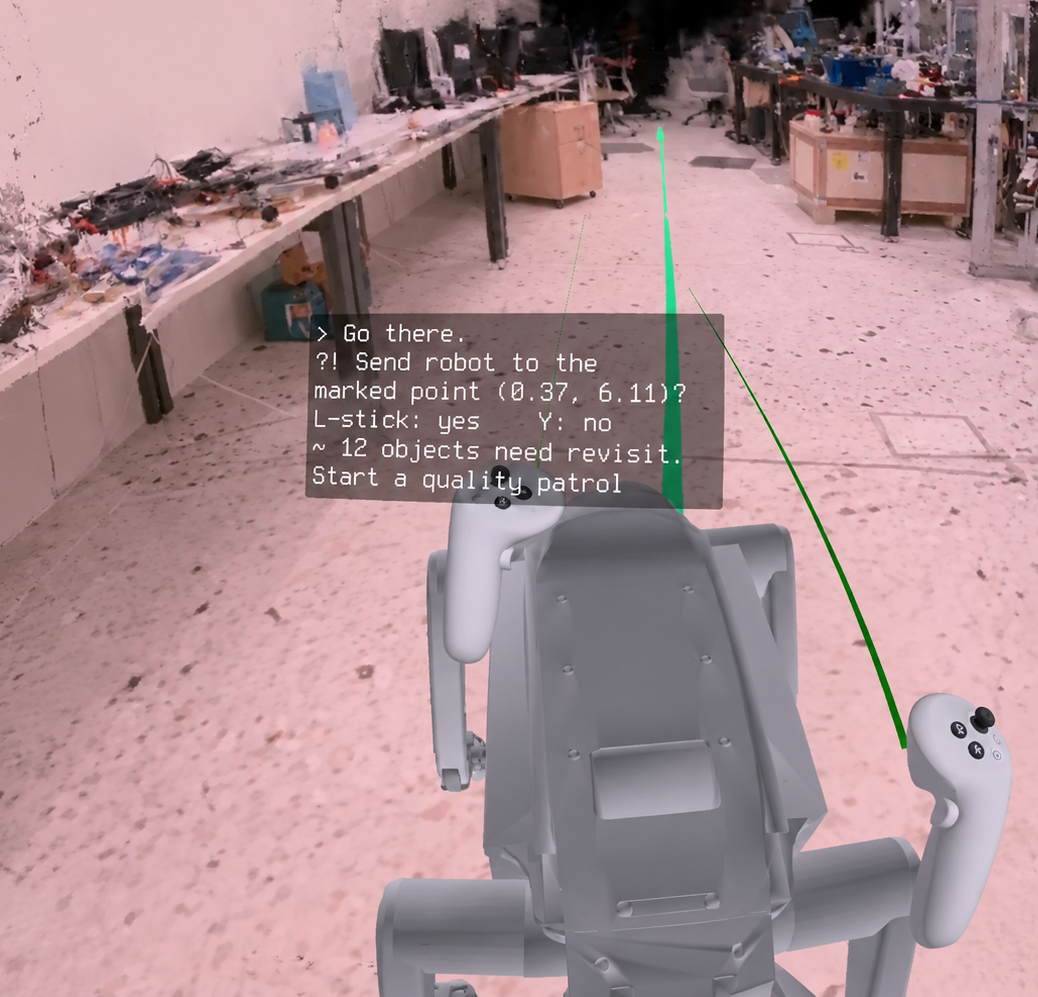}
     \caption{Example of operator interaction in the VR client. The controller ray provides spatial grounding for a robot navigation request, which is presented to the operator for confirmation before execution.}
     \vspace{-4mm}
     \label{fig:vr_interaction}
 \end{figure}

Requests are processed by a two-layer routing architecture. A deterministic pre-router handles unambiguous requests and policy checks. Remaining requests are processed by a schema-constrained language-model router. In the deployed system, this role is served by a local Qwen3-VL-8B model~\cite{qwen3vl}, which receives the transcript, relevant dialogue history, spatial grounding, and a compact serialization of the current Scene Model. The model is constrained to produce a typed JSON decision $\mathbf{d}$ containing a tool or mission name and structured arguments. Post-validation then checks argument completeness, object existence, merge-resolved IDs, and state consistency before any effect is executed. Visual requests additionally provide a rendered scene or VR viewport image to the same multimodal model. Robot-directed actions are staged for operator confirmation, while multi-user state is synchronized through shared selections and annotations.

Figure~\ref{fig:vr_interaction} shows ray-grounded navigation with operator confirmation in the VR client.

\subsection{Agent-robot interaction}

Agent-robot interaction separates high-level goals from robot-side execution. Using scene geometry, robot pose, and mapping quality, the agent requests navigation or additional observations. Goals may refer to objects, scene coordinates, or predefined regions.

The robot backend executes predefined missions, such as object approach, inspection, reinspection, and patrol. Navigation goals require operator approval before execution. The agent tracks the reported outcome, distinguishing arrival, failure, and timeout.

Observation requests remain separate from their execution: registering a revisit does not confirm that the robot moved or collected new data. These outcomes must be verified through execution feedback and incoming sensor observations.

\section{EVALUATION AND EXPERIMENTS}

\subsection{Experimental Setup}

\begin{figure}[t]
    \vspace{3pt}
    \centering
    \includegraphics[width=\linewidth]
        {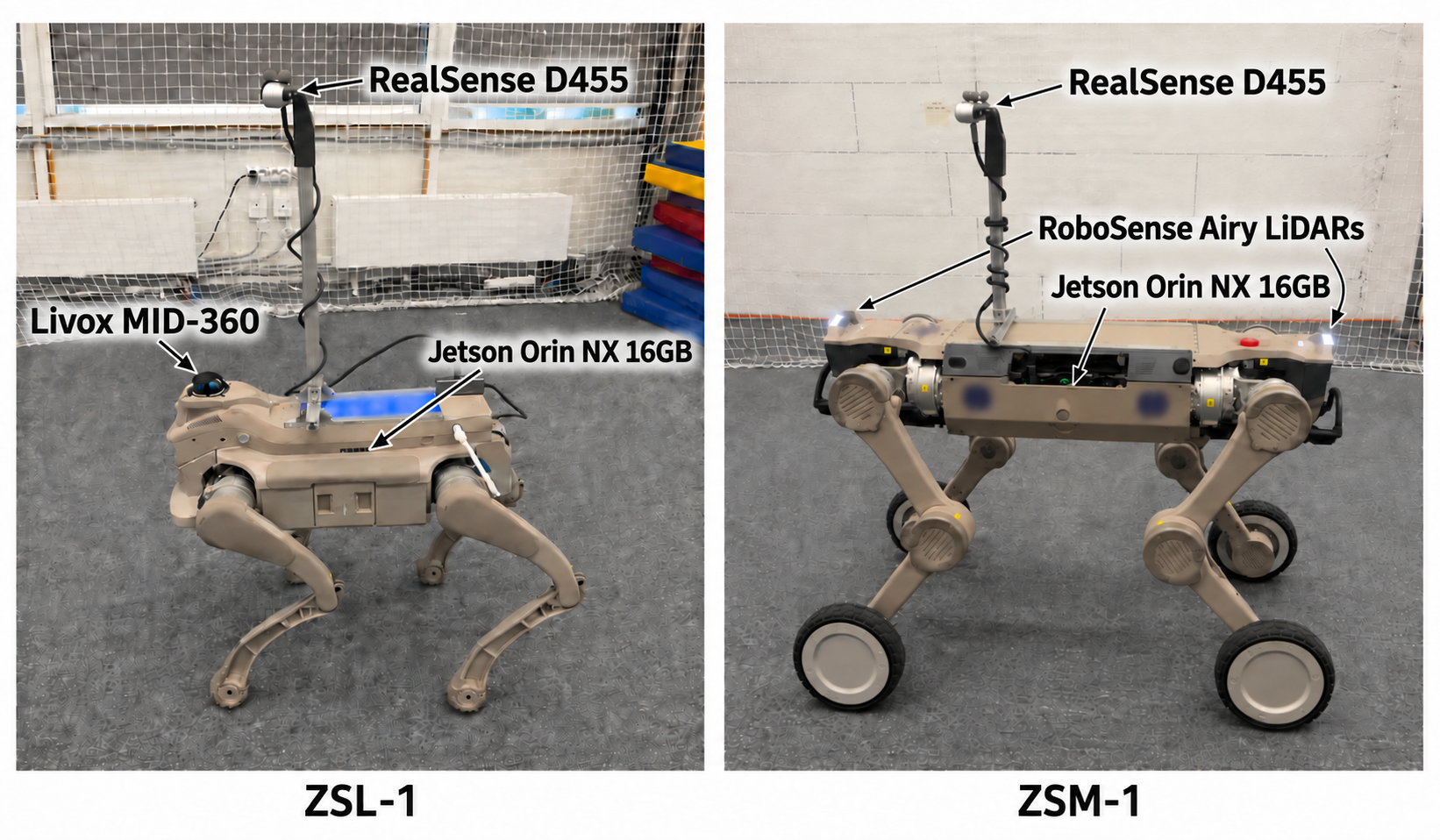}
    \caption{Quadruped robotic platforms used in this work: ZSL-1 (left) and ZSM-1 (right), with mast-mounted RGB-D cameras.}
    \vspace{-4mm}
\label{fig:robot_platforms}
    \label{fig:robot_hardware}
\end{figure}

To evaluate our semantic reconstruction, VR rendering, Human-Agent and Agent-Robot Interaction subsystems, we conducted a series of real-world experiments.

The reconstruction was conducted on a desktop computer equipped with an Intel Core i9 14900K CPU and an NVIDIA GeForce RTX 5090 GPU with 32~GB of video
memory, with a second identical computer serving as a semantic server, hosting SAM3 and a local VLM. Separating the semantics is not a necessary condition for the operation of our system; however, it allows a higher SAM3 segmentation rate and larger reconstructed scenes.

The mobile robot setup comprises two quadrupeds ZSL-1 and an ZSM-1 (Fig.~\ref{fig:robot_hardware}), each equipped with
an NVIDIA Jetson Orin NX 16GB and an upper-mounted RealSense D455. The ZSL-1 uses a Livox MID-360, while the M1 has front- and rear-mounted RoboSense Airy LiDARs; the evaluated M1 configuration uses the front LiDAR and its IMU for SLAM. GLIM localization and Nav2 navigation run onboard under ROS 2 Humble, while reconstruction and agent inference run on an external GPU workstation. RGB-D acquisition is configured
at $640\times480$ and 30 Hz, with LiDAR and IMU data at 10 and 200 Hz. Camera poses combine time-aligned
GLIM estimates with provisional rig extrinsics; separate RGB/depth
intrinsics and factory depth-to-color calibration are retained.
Navigation requires operator approval and remains subject to onboard safety checks, with velocity limits of 0.40 m/s and 0.45 rad/s.

 The elevated viewpoint
provides better visibility of tabletops and other raised
surfaces, supporting their exploration and making the
reconstructed scene easier for human operators to interpret.

The UR10 robotic arm was used for experiments on a workbench-scale scene reconstruction. Multiple (30+) objects were placed on a workbench on which the UR10 was mounted, alongside a checkerboard for hand-eye calibration.

For experiments on humanoid robots, we utilized Unitree G1 and UBTECH Walker TienKung, each equipped with a RealSense D455 RGB-D camera mounted on its head. The only source of camera poses in these experiments was the inline ORB-SLAM3 tracker. Additionally, we tested reconstruction, semantic segmentation, and agent scene tools on a dataset recorded with a handheld D455 camera to explore additional scene-exploration scenarios.

Finally, we conducted experiments with a drone equipped with a D455 camera and Vicon markers for localization. The drone explored a room equipped with Vicon beacons, allowing for precise pose estimation. In the experiments, we compare reconstruction quality using Vicon and ORB-SLAM3 on the recorded dataset.

The agent evaluation is reported separately from the reconstruction and VR measurements. It uses recorded scene states and an isolated in-process sandbox to evaluate how the agent converts human-conceived requests into typed operations. The command corpus has mixed construction: people defined the task intents and command content, while part of the phrasing set consists of controlled variants of those human-defined tasks. The primary routing block evaluates the deployed local Qwen3-VL-8B router and a rules-only baseline on 1,471 cases derived from 240 task specifications in 12 families and four recorded scenes. All planned cases, including transport errors and context overflows, remain in the denominator. Additional tests replay recorded object-merge events, evaluate referring expressions against Replica~\cite{replica} ground truth, control the metadata supplied to the VLM, and measure context and shared-GPU effects.

These agent experiments complement the reconstruction, VR, and live robot evaluations reported elsewhere in the paper. Because the sandbox isolates the agent and map interface, its results should be interpreted alongside, rather than as a replacement for, the end-to-end physical system evaluations.

\subsection{Reconstruction Evaluation}

\begin{table*}[t]
\vspace{4pt}
\centering
\footnotesize
\setlength{\tabcolsep}{4pt}
\caption{Online mapping on the same sequences with matched settings.}
\label{tab:gps_slam}
\begin{tabular}{l cc cc cc cc cc cc}
\toprule
& \multicolumn{2}{c}{LAB1} & \multicolumn{2}{c}{Quadruped A$^{a}$} & \multicolumn{2}{c}{UR10$^{b}$} & \multicolumn{2}{c}{Drone bag3} & \multicolumn{2}{c}{Drone new1} & \multicolumn{2}{c}{Tien Kung} \\
& \multicolumn{2}{c}{handheld, 720p} & \multicolumn{2}{c}{ORB-SLAM3, 720p} & \multicolumn{2}{c}{robot poses, 720p} & \multicolumn{2}{c}{Vicon, VGA} & \multicolumn{2}{c}{Vicon, VGA} & \multicolumn{2}{c}{head cam, 720p} \\
\cmidrule(lr){2-3}\cmidrule(lr){4-5}\cmidrule(lr){6-7}\cmidrule(lr){8-9}\cmidrule(lr){10-11}\cmidrule(lr){12-13}
& GPS & CR & GPS & CR & GPS & CR & GPS & CR & GPS & CR & GPS & CR \\
\midrule
PSNR$\uparrow$  & 21.50 & \textbf{22.17} & 22.86 & \textbf{26.54} & 22.83 & \textbf{24.23} & 16.11 & \textbf{21.91} & 11.44 & \textbf{19.54} & 15.32 & \textbf{23.20} \\
SSIM$\uparrow$  & 0.689 & \textbf{0.738} & 0.744 & \textbf{0.839} & 0.839 & \textbf{0.843} & 0.557 & \textbf{0.639} & 0.428 & \textbf{0.591} & 0.439 & \textbf{0.738} \\
LPIPS$\downarrow$ & 0.358 & \textbf{0.319} & 0.313 & \textbf{0.212} & 0.145 & \textbf{0.143} & 0.417 & \textbf{0.362} & 0.496 & \textbf{0.407} & 0.456 & \textbf{0.253} \\
FPS             & 64.8 & 65.4 & 97.1 & 51.4 & 10.9 & 29.4 & 178 & 158 & 175 & 132 & 98.9 & 66.1 \\
\bottomrule
\vspace{-4mm}
\end{tabular}
\end{table*}

Table~\ref{tab:gps_slam} compares reconstruction quality with the Gaussian-plus-SDF baseline (GPS-SLAM) on the same frames, with matched mapping parameters and, where external poses exist, the same poses. On the manipulator, kinematic poses with time-offset compensation give 23.66\,dB in the quality setting and 24.23\,dB in the boost setting, against 22.44 and 22.83\,dB for the baseline; matching its optimization budget per second of video adds only 0.4\,dB to the baseline, so the gain is not a matter of parameters. On the handheld laboratory sequence CognitiveReality reaches 22.17\,dB with inline visual tracking against 21.50\,dB for the baseline with ICP. On the 720p quadruped sequence~A the baseline's ICP fails once the sequence is subsampled to fit in RAM, so the first 1440 frames are compared at full rate: 26.54 against 22.86\,dB. On two quadrotor flights with motion-capture poses CognitiveReality reaches 21.91 and 19.54\,dB against 16.11 and 11.44\,dB, and on the Tien Kung head camera 23.20 against 15.32\,dB within the 8\,m fusion range.

On the ZSL-1 quadruped D455 camera, gyroscope gating raises PSNR from 18.09 to 20.70\,dB and the budget-calibrated schedule to 22.73\,dB, the level of the vibration-free body camera (22.89\,dB); rotation-triggered rounds alone add 0.78-0.83\,dB. Uncertainty-weighted fusion raises the full sequence~A from 26.45 to 27.00\,dB, and to 27.35\,dB with third-order spherical harmonics. Instance semantics do not change rendering: on Replica office0 PSNR is 38.95\,dB with and without them.

CognitiveReality runs above the camera rate on every sequence: 51-66 frames per second on the 720p sequences, 132-158 on the VGA flights and 29-42 on the manipulator at 20\,Hz input. The baseline is faster on most sequences (65-178) because it holds all frames in RAM, runs no visual tracker and skips the noise-weighted fusion; on the manipulator, with the same optimization budget, it sustains 10.9 frames per second on its subsampled 10\,Hz stream.

In Table~\ref{tab:replica_sem} we compare our semantic pipeline in zero-shot 3D semantic segmentation on Replica (8 scenes, 51 classes, ground-truth poses, one semantic frame per 10 input frames) with a baseline, for which we used HoloAgent-Memory~\cite{zhou2026holoagent}. Labels are transferred to the vertices of the semantic mesh (k=5 majority); IoU/Acc are averaged over classes or weighted by class frequency. Open-vocabulary rows start from 11 generic classes and grow the vocabulary with the VLM auditor (one confirmation round, 2\,s interval); labels are mapped to the 51 classes by exact match, a synonym list or an LLM judge. The HoloAgent's numbers were calculated with a closed vocabulary while our system was evaluated with several configurations. 

With the closed vocabulary CR is on par in mIoU (29.5 vs. 29.9) and higher in mAcc and in both frequency-weighted metrics (+3.0, +8.7 and +12.9 points). The gap between the two families of metrics reflects the class profile of the two maps: CR segments the large and medium classes that dominate the vertex count well (wall, ceiling, floor and chair at 73-79 IoU, rug, sofa, door and bin at 62-74), while thin or rare classes such as pillar, shelf, pipe and wall plug are missed entirely and count as zero in the unweighted average. HoloAgent-Memory, which assigns classes from per-instance SigLIP features at query time, is more uniform across classes but less accurate on the surfaces that cover most of the scene.

Semantic rate matters only through lost frames: with a 20 fps stream and 2 Hz segmentation, the result is unchanged (29.2), whereas at 3 Hz requests exceed the segmentation server's 0.3 s per frame on cluttered views and are dropped, costing 2 mIoU points; the mapper itself never drops frames. The vocabulary matters more. Growing it online from 11 generic classes with the VLM auditor loses 12 mIoU with the default prompt, because the auditor never proposes blinds, rugs, screens, vents or small wall fixtures and some of its labels have no counterpart among the 51 classes. A structure-aware inventory prompt (where the general instructions on what to look for are stated without any specific classes) recovers 5 points and already exceeds HoloAgent-Memory in the weighted metrics, but the closed-vocabulary rows remain the like-for-like comparison, since HoloAgent-Memory also classifies against the same 51 names.

\begin{table}[t]
\centering
\caption{Zero-shot 3D semantic segmentation on Replica. IoU/Acc are averaged over classes (m) or weighted by class frequency~(f). CR - CognitiveReality, HoloAgent-Memory - baseline. Best in bold, second best underlined.}
\label{tab:replica_sem} 
\small
\setlength{\tabcolsep}{4pt}
\begin{tabular}{@{}lcccc@{}}
\toprule
Method & mIoU & mAcc & f-mIoU & f-Acc \\
\midrule
HoloAgent-Memory & \textbf{29.93} & 43.60 & 57.00 & 65.39 \\
\midrule
CR, closed vocab., 10\,fps / 1\,Hz & \underline{29.53} & \textbf{46.63} & \textbf{65.73} & \textbf{78.33} \\
CR, closed vocab., 20\,fps / 2\,Hz & 29.22 & \underline{45.54} & \underline{65.71} & \underline{78.25} \\
CR, closed vocab., 30\,fps / 3\,Hz & 27.56 & 43.35 & 62.99 & 76.04 \\
CR, open vocab., default prompt & 17.23 & 26.80 & 57.66 & 73.06 \\
CR, open vocab., tuned prompt & 22.45 & 33.12 & 62.73 & 77.47 \\
\bottomrule
\end{tabular}
\end{table}

\subsection{VR Evaluation}

We evaluate two VR delivery modes on Pico 4 Ultra headset: remote stereo image streaming (RSIS) and simplified scene streaming with local rendering (3SLR). In RSIS, the host renders both eye views from the headset pose and projection matrices and streams the encoded stereo images to the headset. Representative runs produced 54-84 host-rendered fps, with 25-46 ms median request-to-decode latency. As the Gaussian map grows, rendering cost increases on the host, while the headset requires little scene-side computation or memory.

In 3SLR, the host instead streams a TSDF mesh and quantized Gaussian updates, while the headset maintains a bounded local scene and renders from the latest OpenXR pose. After the initial transfer, only added, modified, or removed primitives are transmitted, removing per-view network delivery from the head-motion path. The evaluated local-rendering configuration maintains up to 200k resident Gaussians and reaches 66-85 fps with approximately 9-10 ms $p95$ GPU frame time and up to 36.7 dB PC-HS PSNR. Thus, RSIS preserves the full host representation but remains sensitive to rendering and network latency, whereas 3SLR trades local compute, memory, and bounded scene fidelity for network-independent viewpoint updates. The reported timing values characterize different pipeline stages and are not a direct motion-to-photon comparison.

\subsection{Human-Agent Interaction}

We evaluate routing, complete calls, and sandbox effects for the deployed local
Qwen3-VL-8B agent across 1,471 planned cases. It improves tool exact match from
48.33\% to 81.24\% and full-call correctness from 42.96\% to 72.88\% over
rules-only routing, while reaching 84.70\% effect correctness. The difference
between tool and full-call scores isolates argument and scene-reference errors.
Post-validation adds 3.04 percentage points in paired full-call correctness,
and the complete guard path produces no unsafe proposals in 71 sandbox policy
cases; this last result measures proposal filtering, not physical safety.

Temporal replay evaluates object reference validity under dynamic map changes (Table~\ref{tab:agent_temporal}). With causally available merge evidence, the resolver completes 117 of 167 episodes and correctly redirects 101 absorbed identifiers, confirming that the merge-aware mechanism itself keeps references valid while the map evolves. This same evidence localizes the one remaining boundary of the pipeline: the confirmation gate currently forwards the resolved reference without re-checking it against a later map update, so the 40 pending-confirmation cases keep the reference resolved at proposal time. Reusing the already-available merge evidence to revalidate a pending target at confirmation is therefore a targeted extension rather than a change to the reference-resolution mechanism.

\begin{table}[t]
    \vspace{5pt}
    \centering
    \caption{Natural temporal replay over 167 episodes from 87 recorded source
    events. A redirect is a correct successor identifier, not navigation.}
    \label{tab:agent_temporal}
    \scriptsize
\begin{tabular}{lcccc}
\toprule
Replay condition & Ep. / events & Ep. success & Redirects & Silent wrong \\
\midrule
Natural / frozen & 167 / 87 & 0 / 167 & 0 & 142 \\
Natural / latest & 167 / 87 & 16 / 167 & 0 & 40 \\
Natural / causal & 167 / 87 & 117 / 167 & 101 & 40 \\
\bottomrule
\end{tabular}

\end{table}

Referring tests on two Replica scenes separate object selection from map
geometry. The typed agent reaches 21.5/28.1\% target-ID correctness and
13.6/30.3\% Acc@0.25 on office0/office1. With the ground-truth object list, the
language model reaches 61.6/71.9\%, while a deterministic predicate oracle
reaches 98.3/97.8\%. This progression identifies the Scene Model as the main
opportunity for improved grounding while retaining a measurable language
component. We report ID and overlap separately because online boxes represent
partial observations, whereas Replica boxes cover complete mesh geometry.

Saved-render controls characterize how the VLM combines images and map
metadata (Table~\ref{tab:agent_vlm}). Named-target agreement rises from 24.7\%
with images alone to 95.9\% with image and metadata, while metadata alone
reaches 99.3\%. Under conflicting metadata, the model repeats the supplied
class in 91.8\% of 146 cases and acknowledges the conflict in 18.5\%. The
deployed path is therefore highly map-faithful; these agreement measures should
not be interpreted as independent visual-recognition accuracy.

\begin{table}[t]
    \vspace{5pt}
    \centering
    \caption{VLM metadata controls. Named target measures agreement with the
    map label; injected class is reported under conflicting metadata.}
    \label{tab:agent_vlm}
    \scriptsize
\begin{tabular}{lccc}
\toprule
Condition & Named target & Injected class & Conflict ack. \\
\midrule
Image only & 0.247 & -- & 0.048 \\
Metadata only & 0.993 & -- & 0.027 \\
Image + metadata & 0.959 & -- & 0.007 \\
Wrong metadata & 0.021 & 0.918 & 0.185 \\
Screenshot + metadata & 0.897 & -- & 0.144 \\
\bottomrule
\vspace{-4mm}
\end{tabular}

\end{table}

Compact context preserves 72/100 and 70/100 full calls on matched office0 and
lab1 subsets; whole-map lab1 falls to 42/100 with 68 context overflows. This
supports selective scene serialization rather than removing memory fields,
since execution still uses the complete state. In a shared-GPU pilot, 12
commands/min increases Gaussian round time by 12.7\% and peak VRAM by
3,294~MB, with no frame gaps. A separate 362-request live trace measures
1,422/1,993~ms p50/p90 push-to-talk latency. These resource results establish
the tested operating envelope rather than a general real-time guarantee.

\subsection{Agent-Robot Interaction}

We evaluate the loop from an agent request to robot motion in a dedicated session on two quadrupeds, using the whole system stack: mapping from the robot stream, the local router, the bridge to the robot's navigation gateway and the headset rendering. The operator issues two kinds of requests. A navigation request places a goal by pointing at the floor or names the object under the ray, which must be confirmed by voice or stick for safe operation. It counts as successful when the robot reports arrival at the goal the operator marked. A re-observation request asks the agent to re-inspect an object flagged in the Scene Model as poorly reconstructed; the agent approaches it with the 1.2 m standoff, waits for arrival and reads the object's quality again; it counts as successful when the robot arrives and the rendering quality of the object improves. Every failure is attributed either to CognitiveReality (mapper, agent or bridge) or to the robot's own navigation stack.

 Of 30 navigation requests, 26 were executed to the marked goal. Three failed inside CognitiveReality: the floor point recorded for the goal did not match the point the operator had marked due to floating noisy Gaussians. One failed because the robot's navigation stack crashed while the goal was active; the request was not retried, and the failure lies outside our system. No request produced motion without confirmation, and no goal reached the robot that the operator had not marked. Counting only failures attributable to our pipeline, 26 of 29 requests (0.90) reached the goal; counting all, 26 of 30 (0.87).

 All 20 re-observation requests issued through the agent were executed: the robot approached the flagged object, the mapper fused the new views, and the rendering PSNR of frames covering the object rose by 2-5 dB depending on the frame, with the Scene Model reporting the higher quality at the agent's verification step. This closes the loop of requesting an object revisit on the real robot: the agent selects the object from the quality entries of the live Scene Model, moves the robot, and verifies the gain on the same model.

\balance
\section{CONCLUSION}

CognitiveReality combines online Gaussian-SDF reconstruction, a versioned semantic Scene Model, typed agent tools, and an immersive operator interface. Persistent object references connect language, pointing, visualization, and robot-directed requests while allowing their failure modes to be evaluated separately.

The controlled agent evaluation shows that the deployed local Qwen3-VL-8B router improves tool exact match from 48.33\% to 81.24\% and full-call correctness from 42.96\% to 72.88\% over rules-only routing. Temporal replay further shows that causally available merge evidence correctly redirects 101 absorbed identifiers in 117 of 167 episodes, while confirmation-time revalidation remains a focused extension of the current pipeline. Metadata controls clarify that VLM agreement with the Scene Model should not be interpreted as independent visual recognition.

The same interface was tested in live robot trials across multiple platforms and localization sources. It reached 26 of 30 marked navigation goals and completed all 20 re-observation requests, with the latter closing the loop from quality-aware object selection to additional sensing and verification. No navigation request produced motion without operator confirmation. These results demonstrate that our integrated system can be used for robust online semantic Guassian mapping and environment inspection, and the Scene Model allows for seamless agent integration for more immersive and safe robot teleoperation.  
\addtolength{\textheight}{-12cm}   








\end{document}